\documentclass[letterpaper]{article} 
\usepackage{aaai2026}  
\usepackage{times}  
\usepackage{helvet}  
\usepackage{courier}  
\usepackage[hyphens]{url}  
\usepackage{graphicx} 
\usepackage{natbib}  
\usepackage{caption} 
\usepackage{algorithm}
\usepackage{algorithmic}

\usepackage{newfloat}
\usepackage{listings}
\DeclareCaptionStyle{ruled}{labelfont=normalfont,labelsep=colon,strut=off} 
\floatstyle{ruled}
\newfloat{listing}{tb}{lst}{}
\floatname{listing}{Listing}
\usepackage{amssymb}
\usepackage{amsmath}

\title{Reasoning Knowledge-Gap in Drone Planning via LLM-based Active Elicitation}
\author{
    Zeyu Fang\textsuperscript{\rm 1},
    Beomyeol Yu\textsuperscript{\rm 1},
    Cheng Liu\textsuperscript{\rm 1}, 
    Zeyuan Yang\textsuperscript{\rm 1}, 
    Rongqian Chen\textsuperscript{\rm 1}, 
    Yuxin Lin\textsuperscript{\rm 2}, 
    Mahdi Imani\textsuperscript{\rm 2}, 
    Tian Lan\textsuperscript{\rm 1}
}
\affiliations{
    \textsuperscript{\rm 1}George Washington University,
    Washington, D.C., USA\\
    \textsuperscript{\rm 2}Northeastern University,
    Boston, MA, USA\\
    joey.fang@gwu.edu, yubeomyeol@gwu.edu, chengl@gwu.edu, zeyuan.yang@gwu.edu, rongqianc@gwu.edu, lin.yuxi@northeastern.edu, m.imani@northeastern.edu, tlan@gwu.edu

}

\begin{document}

\maketitle

\begin{abstract}
Human-AI joint planning in Unmanned Aerial Vehicles (UAVs) typically relies on control handover when facing environmental uncertainties, which is often inefficient and cognitively demanding for non-expert operators. To address this, we propose a novel framework that shifts the collaboration paradigm from control takeover to active information elicitation. We introduce the Minimal Information Neuro-Symbolic Tree (MINT), a reasoning mechanism that explicitly structures knowledge gaps regarding obstacles and goals into a queryable format. By leveraging large language models, our system formulates optimal binary queries to resolve specific ambiguities with minimal human interaction. We demonstrate the efficacy of this approach through a comprehensive workflow integrating a vision-language model for perception, voice interfaces, and a low-level UAV control module in both high-fidelity NVIDIA Isaac simulations and real-world deployments. Experimental results show that our method achieves a significant improvement in the success rate for complex search-and-rescue tasks while significantly reducing the frequency of human interaction compared to exhaustive querying baselines.
\end{abstract}


\section{Introduction}

Human-AI joint planning in the field of Unmanned Aerial Vehicles (UAVs) or drone control has emerged as a critical area of research in intelligent robotics \cite{sapkota2025uavs, lim2021adaptive, papyan2024ai, jain2024unmanned}. Traditionally, human-robot collaboration in open-world environments has been predominantly characterized by unidirectional control handover. In such paradigms, when a UAV encounters uncertainties—such as unidentifiable obstacles or ambiguous environmental semantics—the standard fallback mechanism is to suspend autonomy and transfer control to a professional human operator \cite{luo2023human, wu2023toward, igbinedion2024learning}. However, this approach relies heavily on the assumption that the human supervisor always possesses the immediate situational awareness and expert control skills required to execute the optimal maneuver. In real-world cooperation applications, such as search and rescue (SAR) or exploration missions \cite{schwalb2022study, fang2023implementing}, this assumption often fails: the human co-operator may understand the semantic context of the surrounding environment but may not be proficient in providing the precise low-level manipulations required for optimal planning. Consequently, relying solely on control takeover is often inefficient and imposes a heavy cognitive load.

In this context, we argue that the planning difficulties of modern AI-driven drones often stem not from a lack of control policies but from specific \textit{knowledge gaps}—missing information regarding certain objects in the environment. 
Therefore, a more effective Human-AI joint planning strategy shifts the focus from ``acting'' to ``knowing''~\cite{lai2021towards, chen2025neurosymbolic}; the human supports the autonomous planner by bridging these specific information deficits. However, this introduces a new challenge: the agent must precisely identify the source of uncertainty, convert it into natural language, and perform active elicitation. Besides, to maintain operational efficiency, the AI agent must avoid generic open-ended questions and instead formulate targeted queries that resolve the ambiguity with minimal interaction.

With the recent advancements in Large Language Models (LLMs) such as ChatGPT, a new paradigm for addressing these challenges has become possible \cite{li2026acdzero}. In this paper, we propose a novel human-AI interaction framework for object-driven knowledge-gap reasoning and active elicitation. Unlike standard approaches that treat uncertainty as a monolithic block to be avoided, our method explicitly structures uncertainty using a \textit{Minimal Information Neuro-Symbolic Tree (MINT)}. This framework first constructs a symbolic tree to analyze the source of the knowledge gap. Subsequently, an LLM traverses this structure to curate a binary (Yes/No) query that maximizes information gain regarding the optimal plan. Upon receiving the human's response, the agent updates its internal belief state, prunes the uncertainty tree, and revises its plan accordingly. This process iterates until the decision-making uncertainty is resolved or no further relevant information can be provided.

\begin{figure*}
    \centering
    \includegraphics[width=0.9\linewidth]{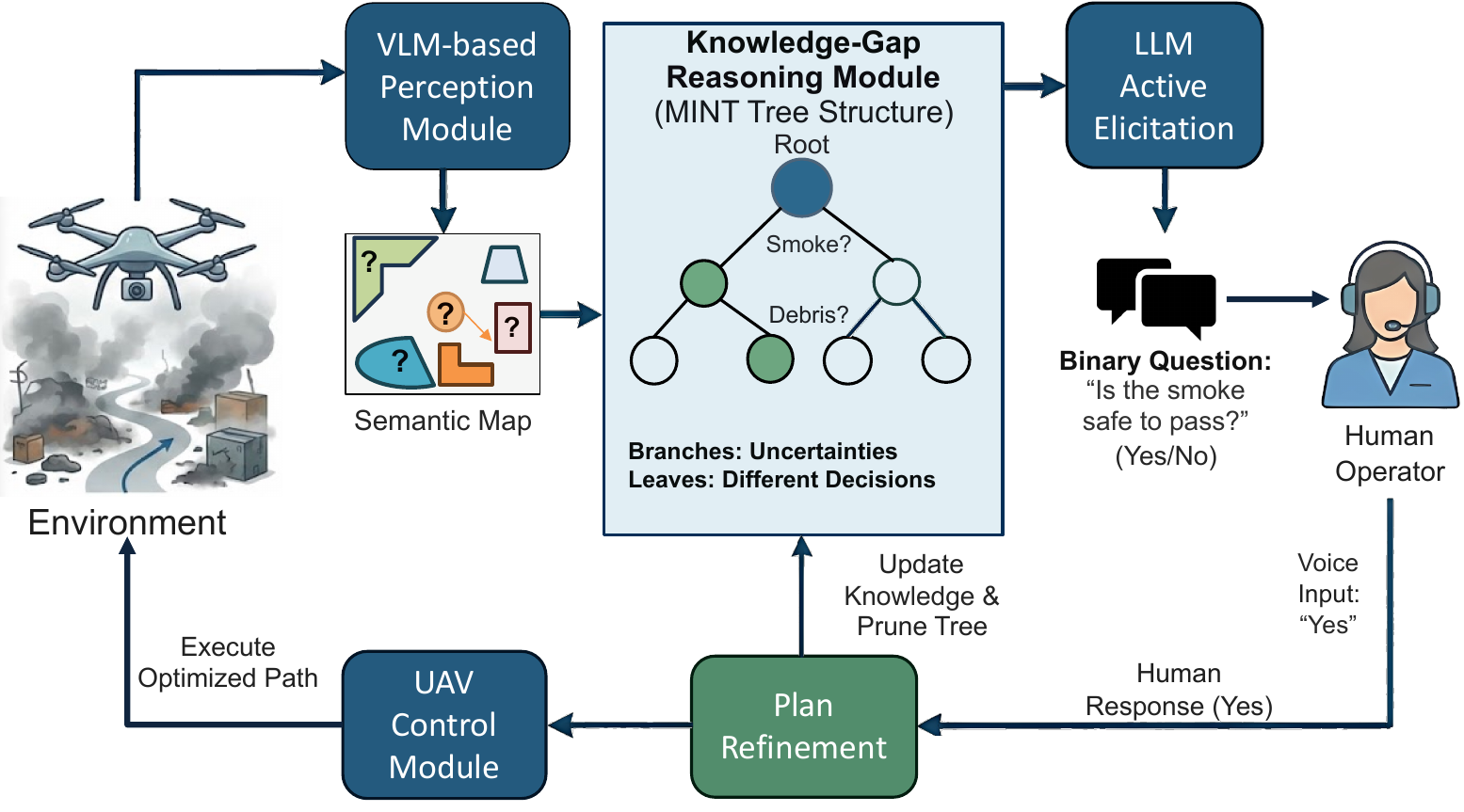}
    \caption{The complete workflow of our methods.}
    \label{fig:workflow}
\end{figure*}

To validate our approach, we developed a comprehensive system workflow, as illustrated in Figure \ref{fig:workflow}, and evaluated it in both high-fidelity simulation and real-world scenarios. Our system integrates a human voice interface, a Vision-Language Model (VLM) for semantic perception, a neuro-symbolic reasoning module, and a low-level UAV controller. By deploying this workflow in complex search-and-rescue tasks involving visual obstructions and unknown hazards, we demonstrate that our method not only successfully completes planning objectives where baselines fail but also significantly reduces the frequency of human interactions required to achieve near-expert performance.

\section{Methodology}

We propose a neuro-symbolic framework for active human-UAV collaboration. Our system enables a planning agent inside the UAV to autonomously identify perceptual or semantic ambiguities during navigation tasks and actively query a human operator to resolve them. The proposed method implements joint-planning and uncertainty mitigation by: (1) Extracting object attributes and identifying potential uncertainties; (2) Knowledge-gap reasoning via MINT, which models the branching possibilities of navigation outcomes; and (3) LLM-driven active elicitation, which formulates optimal binary queries to prune and update the neural-symbolic tree to resolve the uncertainty.

\subsection{Object-Driven Uncertainty Identification}
The UAV is equipped with an RGB-D camera and receives a natural language instruction $I$ (e.g., ``Pick up medicine from the box and bring it to the injured person''). We utilize a VLM to process the visual input $V$ and construct a local semantic map $\mathcal{M}$. The VLM detects objects $O = \{o_1, o_2, ..., o_n\}$ and extracts their attributes (color, shape, semantic class). Normally, for navigation tasks, the system flags \textit{uncertainty} in two forms:
\begin{itemize}
    \item \textbf{Obstacle-based Ambiguity:} When an object's property is crucial for the task but cannot be determined (e.g., whether the smoke in the region is safe to pass).
    \item \textbf{Target-based Ambiguity:} When multiple objects match the description in $I$ (e.g., two distinct objects with the same label "box" exist), which creates high entropy in the sub-goal distribution.
\end{itemize}

Let $u$ denote the set of identified knowledge gaps. If $u = \emptyset$, the UAV proceeds with standard deterministic path-planning via classic algorithms such as A*. Otherwise, if $u \neq \emptyset$, the system triggers the reasoning module to resolve the uncertainty first.

\subsection{Knowledge-Gap Reasoning}
To resolve the identified uncertainty $u$, we constructed MINT to analyze its possible impacts. The uncertainty is evaluated based on its entropy and trajectory divergence to measure its significance and impacts on decision-making. If the control policy is based on reinforcement learning, then the Q-value can also be utilized as a metric \cite{fang2026mint}.

\subsubsection{Tree Construction}
The root node represents the current state $s_t$ with the unresolved knowledge gap $u$. We expand the tree by generating hypotheses $h \in \mathcal{H}$ regarding the unknown variable. For example, if $u$ represents an unknown obstacle property (e.g., smoke), the branches may represent $h_{safe}$ (passable) and $h_{danger}$ (impassable).

For each hypothesis branch, we instantiate a temporary semantic map $\mathcal{M}_h$ and employ a hierarchical planner to generate a corresponding sub-task trajectory $\tau_h$. The cost of the trajectory $C(\tau_h)$ and deviations between two trajectories $d(\tau_{h_1}|\tau_{h_2})$ serve as the evaluation metrics.

\subsubsection{Evaluation Metrics}
We define the impact of a knowledge gap by its effect on the structure of the navigation plan:
\begin{enumerate}
    \item \textbf{Trajectory Divergence:} For obstacle-based uncertainty, we measure the difference between the optimal paths under different hypotheses. If $| C(\tau_{safe}) - C(\tau_{danger})| \leq \delta_c$ and $d(\tau_{safe}|\tau_{danger}) \leq \delta_d$, the uncertainty is irrelevant to the mission, and no query is needed. Here, the $\delta_c$ and $\delta_d$ are two thresholds acting as hyperparameters.
    \item \textbf{Goal Entropy:} For target-based uncertainty (e.g., "Which box to take, black or blue?"), we calculate the Shannon entropy of the potential target distribution $P(g|I, V_t)$. A high entropy indicates a need for clarification.
\end{enumerate}

These metrics are used to construct the neural-symbolic tree from the root node. If one of the metrics is significant (e.g., if one requires a long detour), then the gap is considered critical to cause a difference in decision, and the current node will further branch into two sub-trees with different hypotheses. Otherwise, the current node will be considered a leaf node, and the current planned trajectory will be carried out without further reasoning.

\subsection{Active Elicitation and Plan Refinement}
Once the tree captures the potential future states derived from $u$, a LLM  acts as the inference engine to select the optimal query and convert it into natural language.

\subsubsection{Query Generation}
The LLM analyzes the MINT structure to formulate a binary question $q$ (Yes/No) that maximizes the Information Gain (IG) relative to the planner's objective. The goal is to find a cut in the tree that collapses the hypothesis space $\mathcal{H}$ into a single confident branch.
\begin{equation}
    q^* = \arg \max_{q} \left( H(\mathcal{T}) - \mathbb{E}_{y \in \{yes, no\}} [H(\mathcal{T} | y)] \right)
\end{equation}
where $H(\mathcal{T})$ represents the uncertainty of the current tree state (measured via goal entropy or path variance). The generated question translates technical uncertainties into natural language, such as ``Is the smoke ahead safe to fly through?'' or ``Are you referring to the red box?''.

\subsubsection{Tree Pruning and Execution}
Upon receiving the human operator's binary response $y$, the system prunes all branches inconsistent with $y$. The semantic map $\mathcal{M}$ is updated with the new ``expert'' knowledge (e.g., marking the smoke region as a non-traversable obstacle). The planner then generates a final, optimized trajectory $\tau^*$ based on the resolved map, ensuring safe and efficient navigation.

\section{Experimental Results and Discussion}

We evaluate our proposed framework in two distinct settings: a high-fidelity simulation environment (NVIDIA Isaac) to quantitatively assess planning performance and a real-world environment to demonstrate the feasibility of the integrated voice-interactive workflow.

\begin{figure*}[t]
    \centering
    \includegraphics[width=\linewidth]{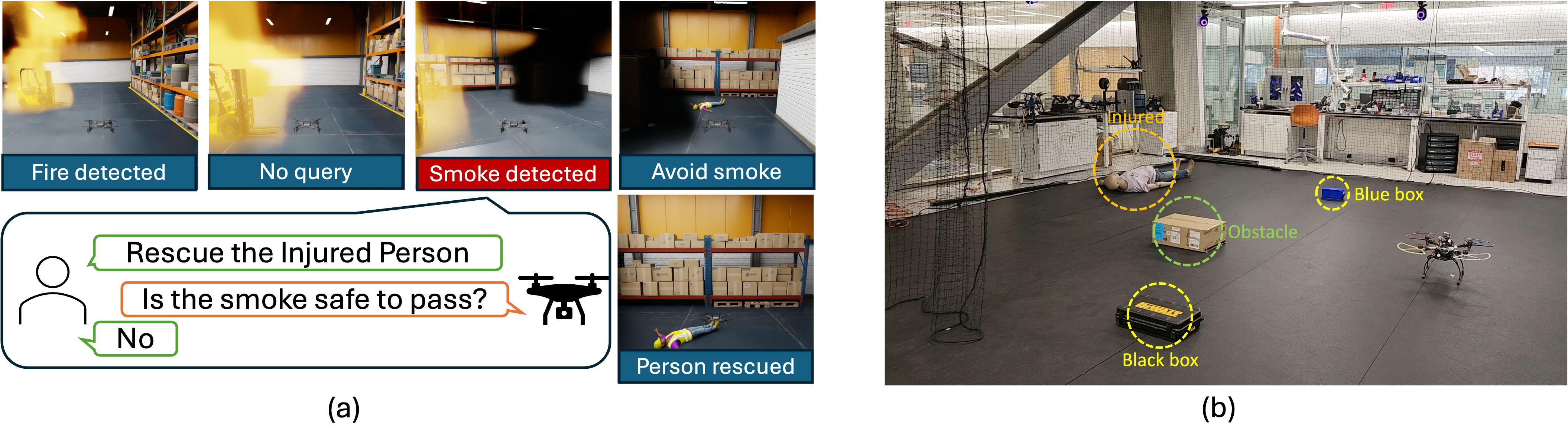}
    \caption{The illustration of the testbeds. (a) The typical planning stage in the NVIDIA Isaac simulation environment. (b) The real-world environment for deployment.}
    \label{fig:env}
\end{figure*}

\subsection{Simulation Experiments: NVIDIA Isaac}
\label{subsec:sim_exp}

We constructed a simulated disaster response scenario in NVIDIA Isaac , where a UAV is tasked with navigating in an on-fire warehouse to locate and rescue an injured person. The environment contains obstacles and is designed with two specific sources of uncertainty ($u$) that affect planning:
\begin{enumerate}
    \item \textbf{Navigation Constraint ($u_{nav}$):} A region covered by volumetric smoke. The UAV cannot visually determine if the smoke is toxic (non-traversable) or safe to fly through (shortcut traversable).
    \item \textbf{Goal Ambiguity ($u_{goal}$):} A closed room that may or may not contain a second potential victim, representing a reward uncertainty.
\end{enumerate}

\subsubsection{Experimental Setup and Baselines}
To validate the effectiveness of our Knowledge-Gap Reasoning module, we varied the spatial configuration of the smoke and the closed room relative to the optimal path. This setup ensures that in some scenarios, the uncertainty is irrelevant to the optimal decision (e.g., the smoke is far from the target), while in others, it is critical.

We compared our method (\textbf{MINT}) against two baselines:
\begin{itemize}
    \item \textbf{Pure LLM (Passive):} The agent generates a plan using standard $A^*$ logic based on a VLM description without the ability to query. It treats all uncertain regions as obstacles (conservative) or free space (risky), leading to either suboptimal detours or collisions.
    \item \textbf{Exhaustive Query (Always Ask):} The agent triggers a human interaction for every detected uncertainty (smoke or closed doors) regardless of its impact on the current trajectory.
\end{itemize}

\subsubsection{Results Analysis}
The results are summarized in Table \ref{tab:sim_results}. Our approach demonstrated a significant improvement in efficiency. In scenarios where the smoke was not on the critical path, MINT successfully pruned the uncertainty tree and avoided issuing unnecessary queries, whereas the \textit{Exhaustive Query} baseline incurred a penalty for every uncertainty.

Compared to the \textit{Pure LLM} baseline, which achieved a success rate of only $\mathbf{77}\%$, our method achieved a success rate of $\mathbf{100}\%$. Furthermore, compared to the \textit{Exhaustive Query} baseline, we reduced the number of human interactions by $\mathbf{30}\%$ while maintaining a comparable task success rate. This confirms that our neuro-symbolic tree effectively filters out "noise" uncertainties that do not affect the decision boundary.

\begin{table}[h]
\centering
\label{tab:sim_results}
\begin{tabular}{lcc}
\hline
\textbf{Method} & \textbf{Success Rate (\%)} & \textbf{Avg. Queries} \\ \hline
Pure LLM & $77.0$ & $0.0$ \\
Exhaustive & $100.0$ & $2.0$ \\
\textbf{MINT (Ours)} & $100.0$ & $1.4$ \\ \hline
\end{tabular}
\caption{Comparison of Success Rate and Interaction Cost in Simulation}
\end{table}

\subsection{Real-World Deployment}
\label{subsec:real_world}

To validate the system in a physical setting, we deployed the complete workflow on a quadrotor platform . The system integrates a VLM-based perception module (for object detection and attribute extraction), a voice interaction interface (Speech-to-Text/Text-to-Speech), and a low-level flight controller.

\subsubsection{Scenario Description}
The real-world task was defined as: \textit{"Pick up the medicine from the box and bring it to the injured person."} The environment contained multiple boxes of varying sizes, shapes, and colors, introducing semantic ambiguity regarding the correct target object. The UAV utilized onboard cameras to detect objects and, upon encountering multiple potential targets (high goal entropy), initiated a voice query to the human operator (e.g., "Should I take the red box?").

\subsubsection{Performance}
We conducted $\mathbf{20}$ trials with varying arrangements of decoy boxes. The baseline (Passive Planner) often failed to identify the correct medicine box, resulting in a success rate of only $\mathbf{35}\%$. In contrast, our method achieved a success rate of $\mathbf{100}\%$. For future experiments, we will test whether increasing the complexity of the tasks causes performance to decrease.

Crucially, the interaction felt natural to the operators. The latency from the start to completion of planned trajectories was approximately $\mathbf{20.7}$ seconds. By actively eliciting the necessary visual attributes (e.g., color) through voice, the UAV was able to resolve semantic ambiguities that are impossible to solve via geometric planning alone. The real-world experiments confirm that our object-driven elicitation framework is robust to the noise and variability of physical environments.

\section{Conclusion}
In this paper, we present a neuro-symbolic framework for object-driven knowledge-gap reasoning and active elicitation in human-UAV teaming. By explicitly modeling uncertainty through the MINT structure, our approach enables autonomous agents to distinguish between critical and irrelevant knowledge gaps, thereby optimizing the utility of human intervention. We validated the system in both simulated disaster response scenarios and physical real-world tasks, demonstrating that targeted binary queries can resolve semantic ambiguities that traditional geometric planners cannot handle. The results confirm that our method significantly enhances task success rates compared to passive planning baselines while minimizing the interaction burden on the operator. This work establishes a scalable foundation for intuitive, voice-based human-robot collaboration in open-world environments.
Future directions will focus on expanding the neuro-symbolic reasoning capabilities of MINT to handle more complex, non-binary query structures and continuous value elicitation. We also aim to integrate formal methods, such as temporal logic constraints, into the tree construction process to enable safer and more expressive human-robot collaboration in dynamic, open-world environments.

\appendix

\section{Acknowledgments}
This research is based on work supported by the Office of Naval Research under grants N00014-23-1-2850.

\bibliography{aaai2026}

\end{document}